%% file: main.tex
\documentclass[11pt,a4paper]{article}
\usepackage[hyperref]{emnlp2020}
\usepackage{times}
\usepackage{latexsym}
\usepackage{longtable}

\usepackage{microtype}

\PassOptionsToPackage{table}{xcolor}
\usepackage{colortbl}
\usepackage{multirow}
\usepackage{graphicx}
\usepackage{wrapfig}
\usepackage[textsize=tiny]{todonotes}
\usepackage{subcaption}
\usepackage{pifont}
\usepackage{amsmath,amssymb} 
\usepackage{booktabs}
\usepackage[ruled,vlined]{algorithm2e}

\newcommand{\cmark}{\color{blue}{\ding{51}}}

\newcommand{\xmark}{}

\usepackage{enumitem}

\newcommand{\acronym}{\textbf{\texttt{RMM}}}
\newcommand{\acronymnormal}{RMM}
\newcommand{\approach}{Recursive Mental Model}

\newcommand{\Dia}{Dialogue}
\newcommand{\dia}{dialogue}
\newcommand{\dias}{dialogues}

\newcommand{\target}{$t_O$}
\newcommand{\QA}{$QA_{i\text{-}1}$}
\newcommand{\QAfull}{$QA_{1:i\text{-}1}$}

\usepackage{url}

\aclfinalcopy 
\def\aclpaperid{1453} 

\newcommand{\QI}{\includegraphics[width=8pt]{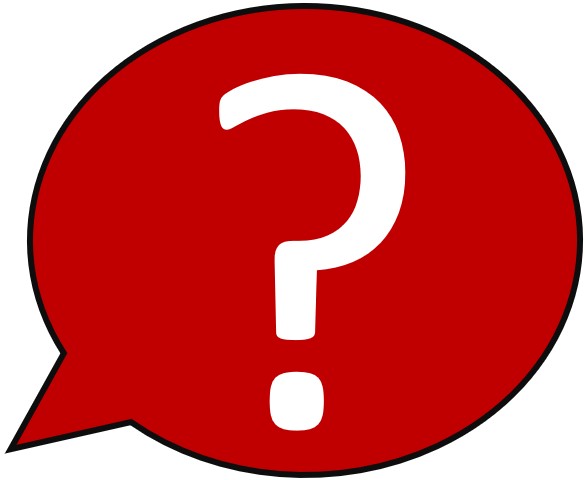}}
\newcommand{\AI}{\includegraphics[width=8pt]{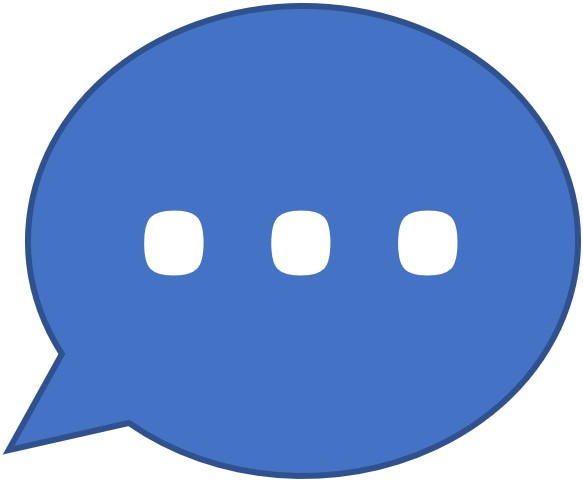}}

\newcommand{\Qr}{$\mathcal{Q}$uestioner}
\newcommand{\Or}{$\mathcal{G}$uide}
\newcommand{\Nr}{$\mathcal{N}$avigator}

\usepackage{float}

\newcommand{\Be}[1]{\textbf{\textcolor{blue}{#1}}}
\newcommand{\PZ}{\phantom{0}}
\newcolumntype{a}{@{\hspace{0pt}}>{\columncolor{gray!15}[0em]}c@{\hspace{0pt}}}

\title{\acronym: A Recursive Mental Model for \Dia{} Navigation}

\author{Homero Roman Roman$^1$ \hspace{1em} Yonatan Bisk$^{1,2}$ \\
   \textbf{Jesse Thomason$^3$ \hspace{1em} Asli Celikyilmaz$^1$ \hspace{1em} Jianfeng Gao$^1$}  \\
  $^1$Microsoft Research\hspace{1em}
  $^2$Carnegie Mellon University \hspace{1em} 
  $^3$University of Washington\\
  {\tt horomanr@microsoft.com \hspace{2em} ybisk@cs.cmu.edu} \\
  }

\date{}

\begin{document}
\maketitle
\begin{abstract}
Language-guided robots must be able to both ask humans questions and understand answers.
Much existing work focuses only on the latter.
In this paper, we go beyond instruction following and introduce a two-agent task where one agent navigates and asks questions that a second, guiding agent answers.
Inspired by theory of mind, we propose the \approach{} (\acronym).
The navigating agent models the guiding agent to simulate answers given candidate generated questions.
The guiding agent in turn models the navigating agent to simulate navigation steps it would take to generate answers.
We use the progress agents make towards the goal as a reinforcement learning reward signal to directly inform not only navigation actions, but also both question and answer generation.
We demonstrate that \acronym{} enables better generalization to novel environments.
Interlocutor modelling may be a way forward for human-agent \dia{} where robots need to both ask and answer questions.
\end{abstract}

\input{01-introduction}
\input{02-related}
\input{03-task}

\input{04-model}
\input{05-results}

\input{06-analysis}
\input{07-conclusions}

\section*{Acknowledgements}
We thank the anonymous reviewers and the AC for the questions they raised and their helpful commentary, which strengthened the presentation of our task and models.
This work was funded in part by ARO grant (W911NF-16-1-0121).

\bibliography{main}
\bibliographystyle{acl_natbib}

\appendix
\input{Appendix}

\end{document}

%% file: 01-introduction.tex
\section{Introduction}
A key challenge for embodied language is moving beyond instruction following to instruction generation, which can require understanding the listener.
The turn-based \dia{} paradigm raises a myriad of new research questions, from grounded versions of traditional problems like co-reference resolution \cite{visual_dialogue} to explicitly modeling theory of mind in order to consider the listener's ability to understand generated instructions~\cite{Bisk2020}.
In this paper, we develop end-to-end \dia{} agents to navigate photorealistic, indoor scenes to reach goal rooms.
We train agents using the human-human Collaborative Vision-and-\Dia{} Navigation (CVDN)~\cite{cvdn} dataset.
CVDN \dias{} are turn-based, with a \textit{navigator} following \textit{guide} instructions and asking questions when needed.

\newpage
Modeling turn-based \dias{} involves four core challenges:
\begin{enumerate}[noitemsep,nolistsep,label={\bfseries C\arabic*}]
\item \label{ch:ask} A \textit{navigator} deciding when to ask a question.
\item \label{ch:que} Generating \textit{navigator} questions.
\item \label{ch:ans} Generating \textit{guide} question answers.
\item \label{ch:act} Generating \textit{navigator} actions.
\end{enumerate}

Prior work has addressed individual components of turn-based \dia\ modeling.
This work is the first to train \textit{navigator} and \textit{guide} agents to perform end-to-end, collaborative \dia{}s with question generation (\ref{ch:que}), question answering (\ref{ch:ans}), and navigation (\ref{ch:act}) conditioned on \dia{} history.

\begin{figure}[t]
    \centering
    \includegraphics[width=\linewidth]{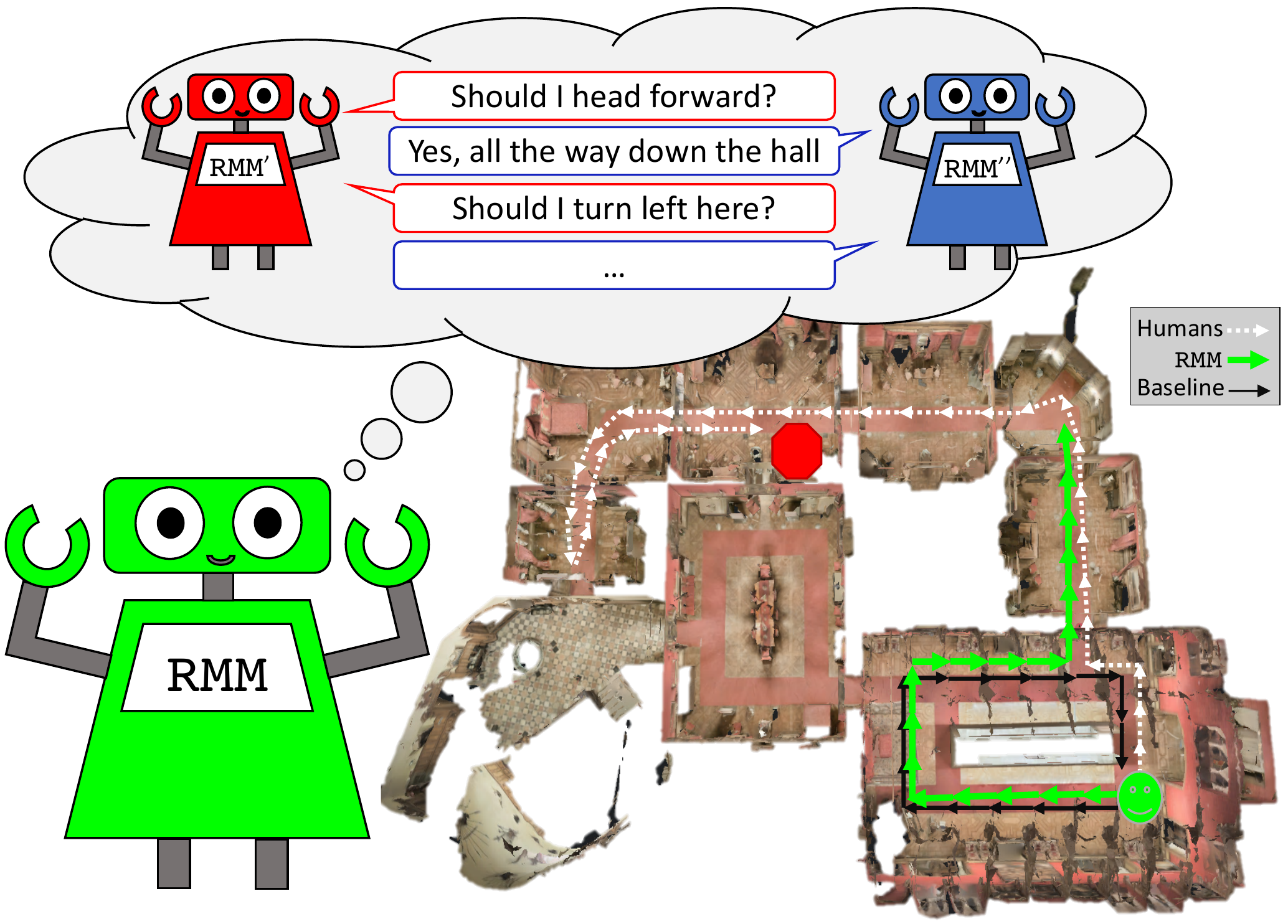}
    \caption{The \acronym{} agent recursively models conversations with instances of itself to choose the right questions to ask (and answers to give) to reach the goal.}
    \label{fig:rmm}
\end{figure}

Theory of mind~\cite{theory_of_mind_child} posits that efficient questions and answers build on a shared world of experiences and referents.
To communicate efficiently, people model both a listener's mental state and the effects of their actions on the world.
Modeling future worlds in navigation~\cite{ghosts} and control~\cite{prospection} are open research questions, and we approximate solutions through a \approach{} (\acronym) of a conversational partner.
Our agent spawns instances of \textit{itself} to simulate the effects of \dia{} acts \textit{before} asking a question or generating an answer to estimate their effects on navigation.
Viewed as a single system, the agents cooperatively search through the space of \dia{}s to efficiently perform embodied navigation.

%% file: 02-related.tex
\section{Related Work and Background}
We build on research in multimodal navigation and the wider literature involving goal oriented \dia.
Table~\ref{tab:dialogue_challenges} summarizes how our work differs from existing work in vision-and-language navigation and task-oriented dialogue modelling.

\paragraph{Instruction Following} tasks an embodied agent with interpreting natural language instructions and visual observations to reach a goal~\cite{jayannavar-etal-2020-learning,self_supervised_imitation_learning,ma2019b,vln,interpret_instructions}.
These instructions describe step-by-step actions the agent needs to take, and can involve the creation of speaker models for data augmentation that provide additional instructions~\cite{speaker-follower}.
This paradigm has been extended to longer trajectories and outdoor environments \cite{vln_streets}, as well as to agents in the real world~\cite{language_to_action, inverse_semantics}.
In this work, we focus on the the simulated, photorealistic indoor environments of the MatterPort dataset \cite{matterport}, and go beyond instruction following alone to a two-agent \dia{} setting.

\paragraph{Navigation \Dia{}s} task a \textit{navigator} and a \textit{guide} to cooperate to find a destination.
Previous work includes substantial information asymmetry between the navigator and guide~\cite{talk_the_walk,narayan-chen-etal-2019-collaborative}.
Information asymmetry can take the form of the \textit{navigator} seeing a bird's eye, abstract semantic map while the \textit{guide} sees egocentric simulation frames~\cite{talk_the_walk}, affecting the kind of dialog possible when low-level visual cues cannot be grounded by the \textit{navigator}.
Other work only investigates the \textit{navigation} portion of the \dia{} without considering text question generation and answering~\cite{cvdn}.
Going beyond models that perform navigation from \dia{} history alone~\cite{Wang:2020,Zhu:2020,Hao:2020}, or decide when to ask \emph{navigator} questions but do so as a simple ``help'' flag with oracle responses~\cite{chi:aaai20,nguyen2019vnla}, in this work we train two agents: a navigator agent that asks questions, and a guide agent that answers those questions.

\begin{table}
\begin{tabular}{@{}p{0.29\textwidth}c@{\hspace{7pt}}c@{\hspace{7pt}}c@{\hspace{7pt}}c@{}}
    \bf Representative Work & \ref{ch:ask} & \ref{ch:que} & \ref{ch:ans} & \ref{ch:act} \\
    \toprule
    \citet{vln} & \xmark & \xmark & \xmark & \cmark \\
    \citet{speaker-follower} & \xmark & \xmark & \cmark & \cmark \\
    \citet{narayan-chen-etal-2019-collaborative} & \xmark & \xmark & \cmark & \xmark \\
    \citet{nguyen2019hanna} & \cmark & \xmark & \cmark & \cmark \\
    \citet{chi:aaai20} & \cmark & \xmark & \xmark & \cmark \\
    \citet{cvdn} & \xmark & \xmark & \xmark & \cmark \\
    \midrule
    RMM & \xmark & \cmark & \cmark & \cmark \\
    \bottomrule
\end{tabular}
\caption{Previous work has addressed subsets of the four key challenges for turn-based navigation \dias{} by training single-turn agents.
No prior work has tackled generating \emph{navigator} questions (\ref{ch:que}); by doing so, our work becomes the first to train two agents jointly on multi-turn \dias{} where agents both produce and consume task-relevant language.
We eschew only the challenge of deciding \textit{when} to ask questions (\ref{ch:ask}), using a fixed heuristic instead.}
\label{tab:dialogue_challenges}
\end{table}

\paragraph{Multimodal \Dia{}} takes several forms.
In Visual \Dia{}~\cite{visual_dialogue}, an agent answers a series of questions about an image that may require \dia{} context.
Reinforcement learning gives strong performance on this task~\cite{visdial_rl}, and such paradigms have been extended to producing multi-domain visual \dia{} agents~\cite{Ju2019AllinOneIC}.
GuessWhat~\cite{Vries2017a} presents a similar paradigm, where agents use visual properties of objects to reason about which referent meets various constraints.
Identifying visual attributes can also lead to emergent communication between pairs of learning agents~\cite{cao2018emergent}.

\paragraph{Goal Oriented \Dia} systems can help a user achieve a predefined goal, from booking flights to learning kitchen tasks \citep{taskorientedgao,2015serbansurvey,bordesweston2017,language_to_action}.
Modeling goal-oriented \dia{} requires skills that go beyond language modeling, such as asking questions to clearly define a user request, querying knowledge bases, and interpreting results from queries as options to complete a transaction.
Many recent task oriented systems are data-driven and trained end-to-end using semi-supervised or transfer learning methods \citep{gpt2dialog,neuralbelieftracker}. 
However, these data-driven approaches may lack grounding between the text and the environment state.
Reinforcement learning-based \dia{} modeling~\cite{rewardshaping,deeprlfordialog,deeprlfordialog1} can improve completion rate and user experience by helping ground conversational data to environments.

%% file: 03-task.tex
\section{Task and Data}

Our work creates a two-agent \dia{} task, building on the CVDN dataset~\cite{cvdn} of human-human \dia s.
In that dataset, a human \Nr{} and \Or{} collaborate to find a goal room containing a target object.
The \Nr{} moves through the environment, and the \Or{} views this navigation until the \Nr{} asks a question in natural language (\ref{ch:ask}, \ref{ch:que}).
Then, the \Or{} can see the next few steps a shortest path planner would take towards the goal, and produces a natural language response (\ref{ch:ans}).
\Dia{} continues until the \Nr{} arrives at the goal (\ref{ch:act}).

\noindent We model this \dia{} between two agents:
\begin{enumerate}[noitemsep,nolistsep]
    \item Questioner ($\mathcal{Q}$) \& Navigator ($\mathcal{N}$)
    \item Guide ($\mathcal{G}$)
\end{enumerate}
We split the first agent into its two roles: question asking (\ref{ch:que}) and navigation (\ref{ch:act}).
As input, the agents receive the same data as their human counterparts in CVDN.
Specifically, both agents (and all three roles) have access to the entire \dia{} and visual navigation histories, in addition to a textual description of the target object (e.g., \textit{a plant}). 
The \Nr{} uses this information to execute on a sequence of actions composed of: \texttt{forward}, \texttt{left}, \texttt{right}, \texttt{look up}, \texttt{look down}, and \texttt{stop}.
The \Qr{} asks for specific guidance from the \Or.
The \Or{} is presented with the navigation and \dia{} histories as well as the next five shortest path steps to the goal, given as a sequence of image observations those steps produce.

Agents are trained on human-human \dia{}s of natural language questions and answers from CVDN.
Individual question-answer exchanges in that dataset are underspecified and rarely provide simple step-by-step instructions like ``straight, straight, right, ...''.
Instead, exchanges rely on assumptions of world knowledge and shared context~\cite{pragmatic_reasoning, logic_and_conversation}, which manifest as instructions rich with visual-linguistic co-references such as \textit{should I go back to the room I just passed or continue on?} 

\begin{algorithm}[t]
\begin{small}
\SetAlgoLined
 $loc = p_0$\;
 $hist = t_0$\;
 $\vec{a} \sim \mathcal{N}(hist)$\;
 $loc, hist = \mathrm{update(}\vec{a}, loc, hist)$\;
 \While{$\vec{a} \neq$ \texttt{STOP} \textbf{and} $len(hist) < 20$}{
  $q \sim \mathcal{Q}(hist, loc)$ \tcp*{Question}
  $\vec{s} = \mathrm{path(}loc, goal\mathrm{, horizon=5)}$ \;
  $o \sim \mathcal{O}(hist, loc, q, \vec{s})$ \tcp*{Answer}
  $hist \leftarrow hist + (q,o)$\;
  \For{$a \in \mathcal{N}(hist)$}{
     $loc \leftarrow loc + a$      \tcp*{Move}
     $hist \leftarrow hist + a$\;
  }
 }
 return $(goal - t_0) - (loc - t_0)$
 \end{small}
 \caption{\Dia{} Navigation}
 \label{alg:dialogue-navigation}
\end{algorithm}

The CVDN release does not provide any baselines or evaluations for the interactive \dia{} setting we present, and instead focuses solely on navigation (\ref{ch:act}).
We use the same metric as that work, ``Goal Progress" in meters---the distance reduction between the \Nr{}'s starting position and ending position with respect to the goal location.

\Dia{} navigation proceeds by iterating through the three roles until either the \Nr{} chooses to stop or a maximum number of turns are played (Algorithm \ref{alg:dialogue-navigation}).
In addition to ``Goal Progress'', we report BLEU scores \cite{papineni-etal-2002-bleu} for evaluating the generation of questions and answers by comparing against human questions and answers. Note, in our \dia{} setting, Goal Progress also implicitly measures the utility of generated language and is therefore complementary to BLEU when evaluating utility versus fluency.

%% file: 04-model.tex
\begin{figure*}[ht]
    \centering
    \begin{subfigure}[t]{0.5\textwidth}
        \centering
        \includegraphics[width=\linewidth]{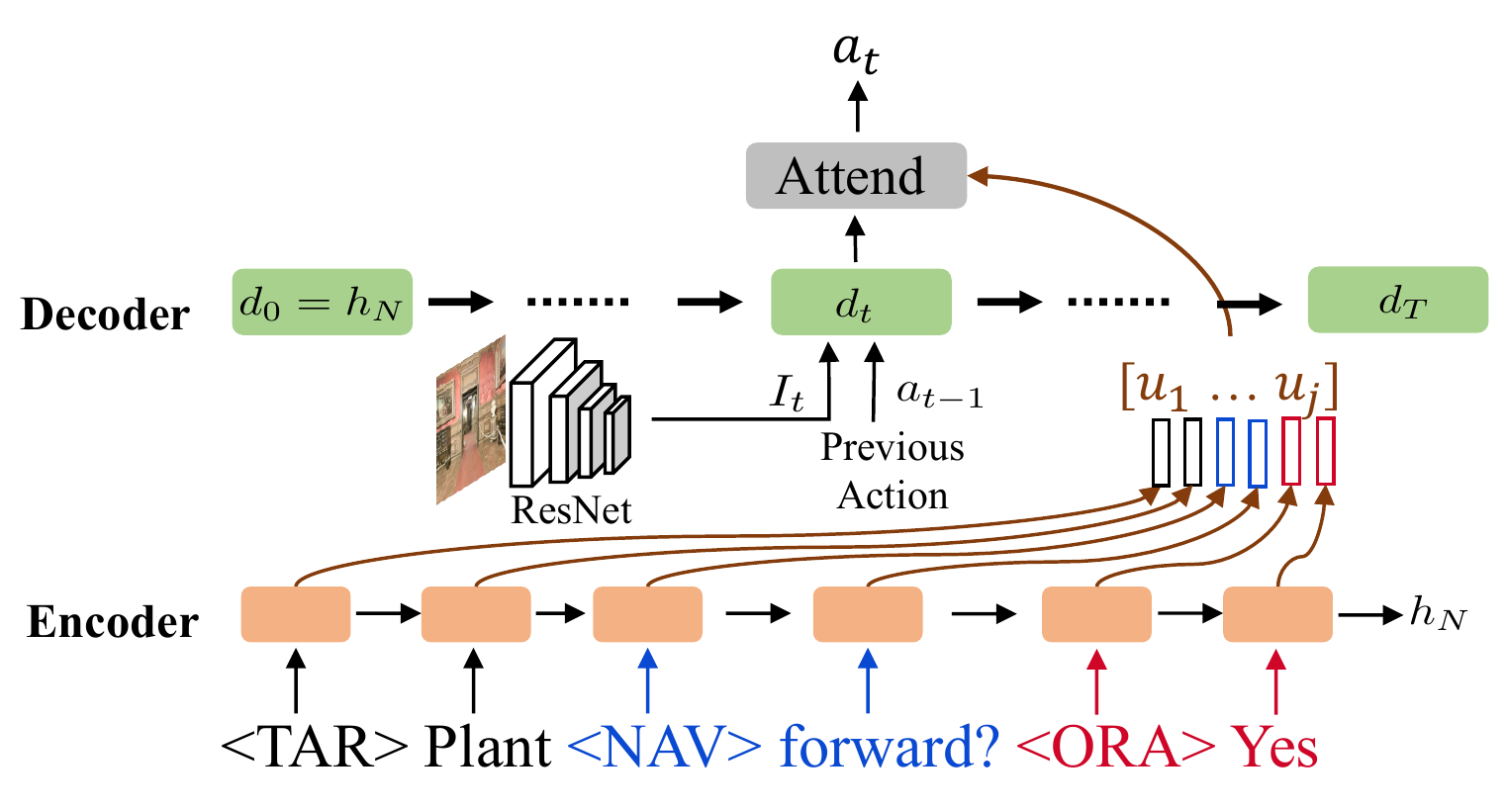}
        \caption{\Dia{} and action histories combined with the current observation are used to predict the next navigation action.}
        \label{fig:action_decoder}
    \end{subfigure}%
    \hspace{1em} 
    \begin{subfigure}[t]{0.4\textwidth}
        \centering
        \includegraphics[width=0.8\linewidth]{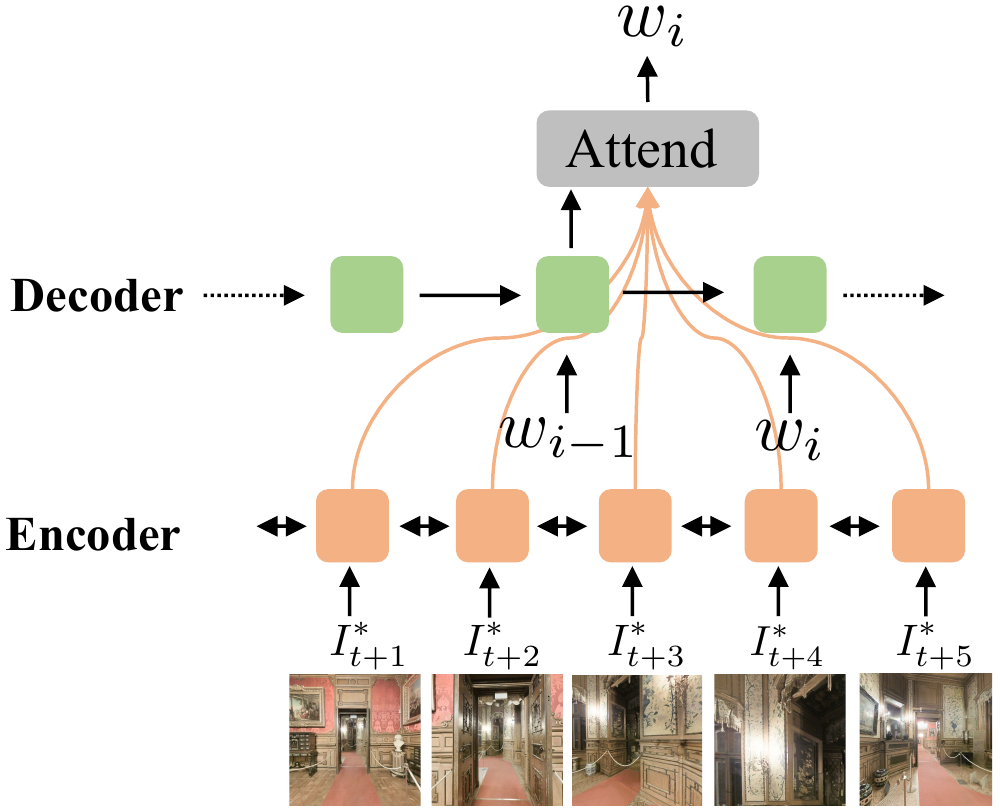}
        \caption{\Or{} Bi-LSTM over the path is attended to during decoding for answer generation.}
        \label{fig:language_decoder}
    \end{subfigure}
        \caption{Our backbone Seq2Seq architectures are provided visual observations and portions of the \dia{} history when taking actions or asking/answering questions.}
        \label{fig:models}
\end{figure*}

\section{Models}
We introduce the \approach{} (\acronym) as an initial approach to our new full \dia{} CVDN task formulation.
Key to this approach is allowing component models (\Nr{}, \Qr{}, and \Or{}) to learn from each other and roll out possible \dia{}s and trajectories.
We compare our model to a traditional sequence-to-sequence baseline, and we explore data augmentation using the Speaker-Follower method~\cite{speaker-follower}.

\subsection{Sequence-to-Sequence Architecture}
The underlying architecture, shown in Figure \ref{fig:models}, is shared across all approaches.
The core \dia{} tasks are navigation action decoding and language generation for asking and answering questions.
We present three sequence-to-sequence \cite{attention_mechanism} models to perform as \Nr, \Qr, and \Or.
The models rely on an LSTM \cite{lstm} encoder for the \dia{} history.
To encode visual observations, our models take the penultimate ResNet~\cite{resnet} layer as the image observation.
Future work may explore different and more nuanced encoding architectures.

\paragraph{Navigation Action Decoding (\ref{ch:act})}
Initially, the \dia{} context is just a target object \target{} category, for example ``plant.''
The goal room contains an instance of that category.
As questions are asked and answered, the \dia{} context grows.
Following prior work~\cite{vln,cvdn}, \dia{} history words $\vec{w}$ words are embedded as 256 dimensional vectors and passed through an LSTM to produce $\vec{u}$ context vectors and a final hidden state $h_N$.
The hidden state $h_N$ is used to initialize the LSTM decoder.
At every timestep the decoder is updated with the previous action $a_{t-1}$ and current image observation $I_{t}$.
The hidden state is used to attend over the language $\vec{u}$ and predict the next action $a_t$ (Figure \ref{fig:action_decoder}).

We pretrain the \Nr{} on the navigation task alone before fine-tuning in the full \dia{} setting that we introduce.
The next action is sampled from the model's predicted logits, and the episode ends when either a \texttt{stop} action is sampled or 80 actions are taken~\cite{cvdn}.  

\paragraph{Speaker Models (\ref{ch:que} \& \ref{ch:ans})}
To generate questions and answers, we train sequence-to-sequence models (Figure \ref{fig:language_decoder}) where an encoder takes in a sequence of images and a decoder produces a sequence of word tokens.
At each decoding timestep, the decoder attends over the input images to predict the next word of the question or answer.
This model is also initialized via training on CVDN \dia{}s.
In particular, question asking (\Qr) encodes the images of the current viewpoint where a question is asked, and then decodes the question tokens produced by the human \Nr.
Question answering (\Or) encodes images of the next five steps the shortest path planner would take towards the goal, then decodes the language tokens produced by the human \Or.
Pretraining initializes the lexical embeddings and attention alignments before fine-tuning in the collaborative, turn-taking setting we introduce in this paper.

\paragraph{Conditioning Context}
We define three levels of \dia{} context given as input to our \Nr{} agents in order to evaluate how well they utilize the generated conversations.
We compare agents' ability to navigate to the goal room given:

\begin{itemize}[align=parleft, labelsep=2em,leftmargin =\dimexpr\labelwidth+\labelsep\relax]
    \item[\target] the target object present in the goal room;
    \item[\QA] additionally the previous question-and-answer exchange;
    \item[\QAfull] additionally the entire \dia{} history.
\end{itemize}

We constrain the \Qr{} and \Or{} speaker models to condition on fixed contexts.
The \Qr{} model 
takes as input the current visual observation $I_t$ and the target object $t_O$.
The \Or{} model 
takes 
the visual observations $I^*_{t+1:t+5}$ of the next five steps of navigation according to a shortest path planner, the target object $t_O$, and the last question $Q_{i-1}$ generated by the \Qr{}.\footnote{This limits phenomena like co-reference, but dramatically reduces the model's input space.  Handling arbitrarily long contexts with limited training data is left to future work.}

\subsection{\approach}
\begin{figure}
\centering
\includegraphics[width=\linewidth]{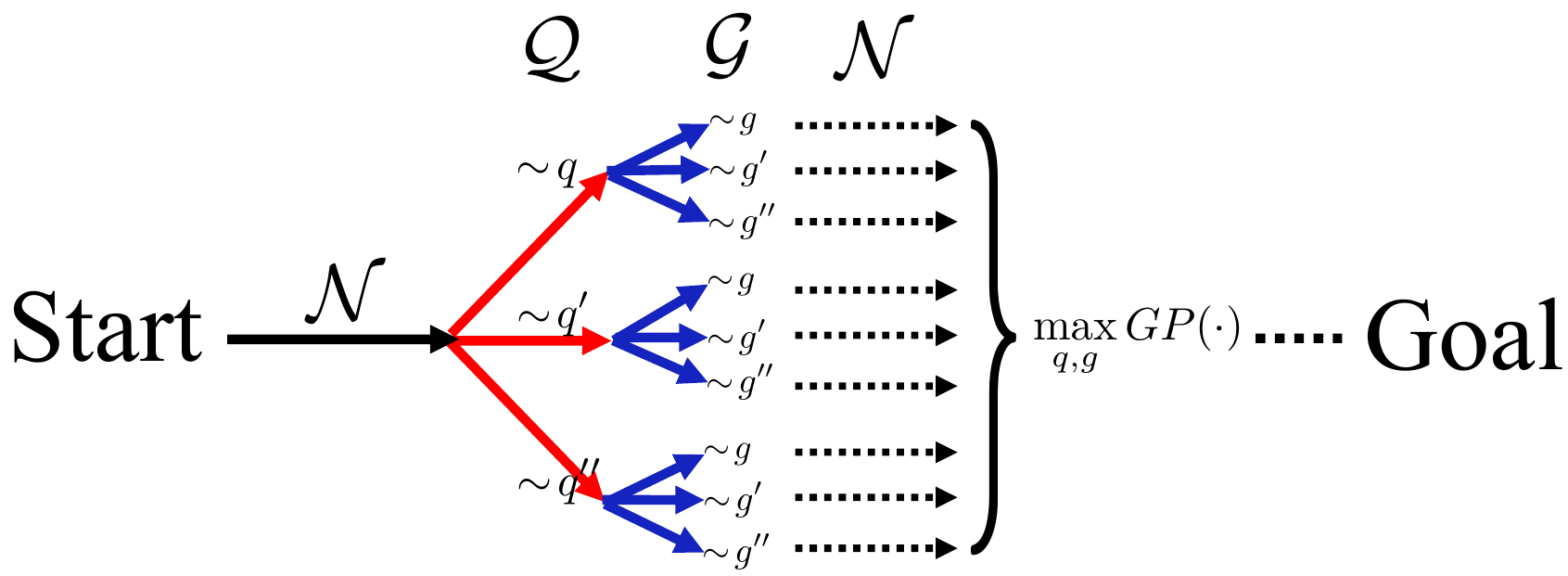}
\caption{The \approach{} allows for each sampled generation to spawn a new \dia{} and corresponding trajectory to the goal.  The \dia{} that leads to the most goal progress is followed by the agent.}
\label{fig:rl_recursive}
\end{figure}

We introduce the \approach{} agent (\acronym),\footnote{\url{https://github.com/HomeroRR/rmm}}
which is trained with reinforcement learning to propagate feedback from navigation error through all three component models: \Nr{}, \Qr{}, and \Or{}.
In this way, the training signal for question generation includes the training signal for answer generation, which in turn is derived from the training signal from navigation error.
The agent's progress towards the goal in the environment informs the \dia{} itself; each model educates the others (Figure \ref{fig:rl_recursive}).

Each model among the \Nr{}, \Qr{}, and \Or{} may sample $N$ trajectories or generations of max length $L$.
These samples in turn are considered recursively by the \acronym{} agent, leading to $N^T$ possible \dia{} trajectories, where $T$ is at most the maximum trajectory length.
To prevent unbounded exponential growth during training, each model is limited to a maximum number of total recursive calls per run.
Search techniques, such as frontiers \cite{Ke2019}, could be employed in future work to guide the agent.

\paragraph{Training}
In the \dia{} task we introduce, the agents begin only knowing the name of the target object.
The \Nr{} agent must move towards the goal room containing the target object, and can ask questions using the \Qr{} model.
The \Or{} agent answers those questions given a privileged view of the next steps in the shortest path to the goal rendered as visual observations.

We define two different loss functions to learn the parameters $\theta$ of the \Nr{}  agent.
We learn a policy $\pi_\theta(\tau|t_O)$ which maximizes the log-likelihood of the shortest path trajectory $\tau$ given target object $t_O$ present in the goal room (Eq. \ref{eq:goal}).
The action decoder $\textbf{a}_t = \textit{f}_{\theta_D}(\textbf{z}_t,I_t)$ takes language encoder $\textbf{z}_t  = \textit{f}_{\theta_E}(w_{1:t})$ as input along with the image observations $I_t$ at time $t$.
\Dia{} context at time $t$, $w_{1:t}$ is input to the language encoder. The cross entropy loss is defined as:
\begin{align}
  J_{CE}(\theta) & = -\sum_{t=1}^{T} log \ \pi_\theta(\textbf{a}_t|I_t, t_O, w_{1:t}) \label{eq:goal}
\end{align}

Our second \Nr{} RL agent loss is standard policy gradient based Advantage Actor Critic~\cite{rl} 
minimizing a k-step TD\footnote{Temporal Difference} error of the critic, $J_{RL}(\theta)$:
\begin{align}
  = & -\ \sum_{t=1}^{T} \textbf{A}^\pi \ log \ \pi_\theta(\textbf{a}_t|I_t, t_O, w_{1:t}) + \frac{1}{2}\sum_{t=1}^{T} (\textbf{A}^\pi)^2 \label{eq:a2c}
\end{align}
 
\noindent $\textbf{A}^\pi$=$r_{t+1}$+$V^\pi(I_{t+1})$-$V^\pi(I_t)$ is the advantage function in Eq. \ref{eq:a2c}, where
$r_{t+1}$ is the reward measured by the goal progress and the $V^\pi$ denotes the state-value (critic) model. The first term in Eq. \ref{eq:a2c} is the actor loss, while the second term is the critic (value) loss of the advantage actor critic loss function. 
The overall system is trained end-to-end using sum of the RL agent loss of the navigator agent $J_{RL}(\theta)$ and the  cross entropy loss between the ground truth and the generated trajectories, $J_{CE}(\theta)$. 
The speaker model parameters are also updated via the sum of the standard question/answer generation cross entropy and the composite \Nr{} agent loss from the branch with the max goal progress. 
 
\paragraph{Inference}
\label{sec:inference}
During training, exact environmental feedback---the remaining distance to the goal---can be used to evaluate samples and trajectories.
This information is not available during inference, so we instead rely on the navigator's confidence to determine which of several sampled paths should be explored.
For every question-answer pair sampled, the agent rolls forward five navigation actions per sequence, and the trajectory sequence with the highest probability is used for the next timestep.
This heuristic does not guarantee that the model is progressing towards the goal, but empirically confidence-based estimation enables progress.

\subsection{\Dia{} Gameplay}
As is common in \dia{} settings, there are several moving pieces and a growing notion of state throughout training and evaluation.
In addition to the \Nr{}, \Qr{}, and \Or{}, the \Nr{} agent also needs to determine when to invoke the \Qr{} model to get supervision from the \Or{} (\ref{ch:ask}).
We leave this component---when to ask questions---for future work and set a fixed number of steps before asking a question.
We invoke the \Qr{} model after every $4$ navigation steps based on the human average of $4.5$ steps between questions in CVDN.

Setting a maximum trajectory length is required due to computational constraints as the the language context $w_{1:j}$ grows.
Following \citet{cvdn}, we use a maximum navigation length of 80 steps, leading to a maximum of $\frac{80}{4} = 20$ question-answer exchanges per \dia{}.

We use a single model for question and answer generation, and indicate the role of spans of text by prepending \texttt{<NAV>} (\Qr{} navigation questions) or \texttt{<ORA>} (\Or{} answers based on oracle views) tags (Figure~\ref{fig:action_decoder}) to condition the generation task.
During roll outs the model is reinitialized to prevent information sharing via the hidden units.

\subsection{Training Details}
\label{ssec:training}

We initialize the \Nr{}, \Qr{}, and \Or{} agents as encoder-decoder LSTM models with 512 hidden dimensions.
The \Nr{} encoder is a forward LSTM, while the \Qr{} and \Or{} speaker models use bi-LSTM encoders.
We use the 512 dimensional penultimate ResNet layer for image observations $I_t$, embed words $w$ in 256 dimensions, and embed actions in $32$ dimensions.
The models observe a word history up to $160$ tokens, and can decode up to $80$ actions per episode.
The value/critic module is a linear layer with relu and dropout on top of the hidden state. 

We optimize the \Nr{} models with the Adam optimizer~\cite{kingma:iclr15} with a learning rate of $0.0001$ with weight decay $0.0005$.
For the \Qr{} and \Or{} models, we use an RMSProp optimizer with learning rate $0.0001$.

Models are pretrained on CVDN data with batches of size $100$ for $20,000$ iterations.
During self-play, models are trained with batches of $10$, for \acronym{} with $N=3$, or $100$ else for $5,000$ iterations.
A dropout rate of $0.5$ is used during all training.
All \Nr{} models are trained using student sampling~\cite{vln}. 
In \acronym$_{N=3}$, one action sequence is produced via argmax decoding, while the other two via sampling (no temperature).  The same is true for language decoding but with a temperature of 0.6.  Exploration of how sampler strategies effect performance is left for future work.

\paragraph{Data Augmentation (DA)}
Navigation agents can benefit from generated language instructions~\cite{speaker-follower}.
We augment the baseline model's navigation training data in a fashion similar to the rollouts of \acronym$_{N=3}$ to create a more direct comparison between the baseline and \acronym.
We choose a CVDN conversation 
and sample three action trajectory rollouts, two by sampling an action at each timestep, and one by taking the argmax action at each timestep.
We evaluate those trajectories' progress towards the conversation goal location and keep the best for augmentation.
We give the visual observations of the chosen path to the pretrained \Qr{} model to produce a relevant instruction.
This trajectory paired with a generated language instruction is added to the training data, and we 
downweight the contributions of these noisier pairs to the overall loss, so $loss = \lambda * \text{generations} + (1-\lambda) * \text{human}$.
The choice of $\lambda$ affects the fluency of the language generated;
we use $\lambda=0.1$.

%% file: 05-results.tex
\begin{table}
\centering
\begin{small}
\begin{tabular}{@{}l@{\hspace{2pt}}l@{\hspace{5pt}}c@{\hspace{2pt}}c@{\hspace{2pt}}a@{\hspace{2pt}}c@{\hspace{7pt}}c@{\hspace{2pt}}a@{}}
 & \textbf{Model} &               \multicolumn{4}{@{}c@{}}{\textbf{Goal Progress (m) $\uparrow$}} & \multicolumn{2}{c}{\textbf{\textbf{BLEU} $\uparrow$}} \\
\toprule
  &  & \begin{scriptsize}\target\end{scriptsize}& \begin{scriptsize}\QA\end{scriptsize}& \begin{scriptsize}\QAfull\end{scriptsize}& \begin{scriptsize}\begin{tabular}{@{}c@{}}+\! Oracle\\Stopping\end{tabular}\end{scriptsize} & \begin{scriptsize}\QA\end{scriptsize}& \begin{scriptsize}\QAfull\end{scriptsize}\\
  \midrule
\multirow{5}{*}{\rotatebox[origin=c]{90}{Val Seen}}  
& Seq2Seq          & \textbf{20.1}& 10.5      & \Be{15.0}     & \textbf{22.9} & 0.9    & 0.8     \\
& Seq2Seq  + DA    &  20.1    & 10.5      & 10.0     & 14.2 & 1.3    & 1.3     \\
& \acronym$_{N=1}$ & 18.7     & 10.0      & 13.3     & 20.4 & 3.3    & 3.0     \\
& \acronym$_{N=3}$ & 18.9     & \textbf{11.5} & 14.0 & 16.8 &\textbf{3.4}&\Be{3.6}\\
&  Shortest Path   &  \multicolumn{4}{@{}c@{}}{----------- 32.8 ----------- }    &   & \cellcolor{white}\\
\midrule
\midrule
\multirow{5}{*}{\rotatebox[origin=c]{90}{Val Unseen}}    
& Seq2Seq          & 6.8     & 4.7      & \PZ4.6      & 6.3& 0.5  & 0.5\\
& Seq2Seq + DA     & 6.8     & 5.6      & \PZ4.4      & 6.5& 1.3  & 1.1\\
& \acronym$_{N=1}$ & 6.1     &\textbf{6.1}  & \PZ5.1      & 6.0 & 2.6  & 2.8\\
& \acronym$_{N=3}$ &\textbf{7.3} & 5.5      & \PZ\Be{5.6} &\textbf{8.9} &\textbf{2.9}&\Be{2.9}\\
& Shortest Path    & \multicolumn{4}{@{}c@{}}{----------- 29.3 ------------}                & & \cellcolor{white} \\
\bottomrule
\end{tabular}
\end{small}
\caption{\Dia{} results on CVDN.
Data Augmentation adds noisy training data for the model.
Goal progress evaluates the quality of the inferred navigation trajectory, while BLEU scores estimate the quality of the generated questions and answers.
Evaluations conditioning on the entire \dia{} history are highlighted in \colorbox{gray!20}{gray} with the best results in \Be{blue}.
}
\label{table:cvdn_results}
\end{table}

\section{Results}
\label{section:results}
In Table \ref{table:cvdn_results} we present \dia{} results for our \acronym{} agent and competitive baselines.
We report two main results and four ablations for \textit{seen} and \textit{unseen} house evaluations; the former are novel \dia{}s in houses seen at training time, while the latter are novel \dia{}s in novel houses.

\paragraph{Full Evaluation}
The full evaluation paradigm conditions navigation on the entire \dia{} history \QAfull{} in addition to the original target object \target.
We present two conditions for \acronym{} ($N=1$ and $N=3$).
Recall that $N$ indicates the number of trajectories (\Nr{}) or generations (\Qr{}, \Or{}) explored in our recursive calls.
$N=1$ corresponds to taking the single maximum prediction while $N=3$ allows the agent to sample alternatives (Section \ref{sec:inference}).
While low, the BLEU scores are better for \acronym{}-based agents across settings.

A challenge for navigation agents is knowing when to \texttt{stop}.
Following previous work~\cite{vln}, we additionally report Oracle Success Rates measuring the best goal progress the agents achieve along the trajectory.

In \textit{unseen} environments, the \acronym{}-based agents make the most progress towards the goal and benefit from exploration at during inference ($N=3$), and this result holds when considering Oracle Success.
In \textit{seen} environments, by contrast, the \acronym{}-based agents perform slightly less well than the baseline sequence-to-sequence models on goal progress.
This effect may be a consequence of \textit{environment bias} in navigation simulations where houses are seen at both training and inference time with overlapping paths~\cite{zhang:ijcai20}.

\paragraph{Ablations}
We also include two simpler results: \target, where the agent is only provided the target object and explores based on this simple goal, and \QA{} where the agent is only provided the previous question-answer pair.
Both of these settings simplify the learning and evaluation by focusing the agent on search and less ambiguous language, respectively.
There are two results to note.

First, given only \target{} the \acronym{} trained model with sampling generalizes best to \textit{unseen} environments.
In this setting, during inference all models have the same limited information, so the RL loss and exploration have better equipped \acronym{} to generalize.

Second, several trends invert between the \textit{seen} and \textit{unseen} scenarios.
Specifically, the simplest model with the least information performs best overall in \textit{seen} houses.
This high performance coupled with weak language appears to indicate the models are learning a different (perhaps search based) strategy rather than how to utilize \dia{}.
In the \QA{} and \QAfull{} settings, the agent generates a question-answer pair before navigating, so the relative strength of the \acronym{} model's communication becomes clear.
We analyze the generated language and navigation behavior of our models.

\begin{figure}[t]
\includegraphics[width=\columnwidth]{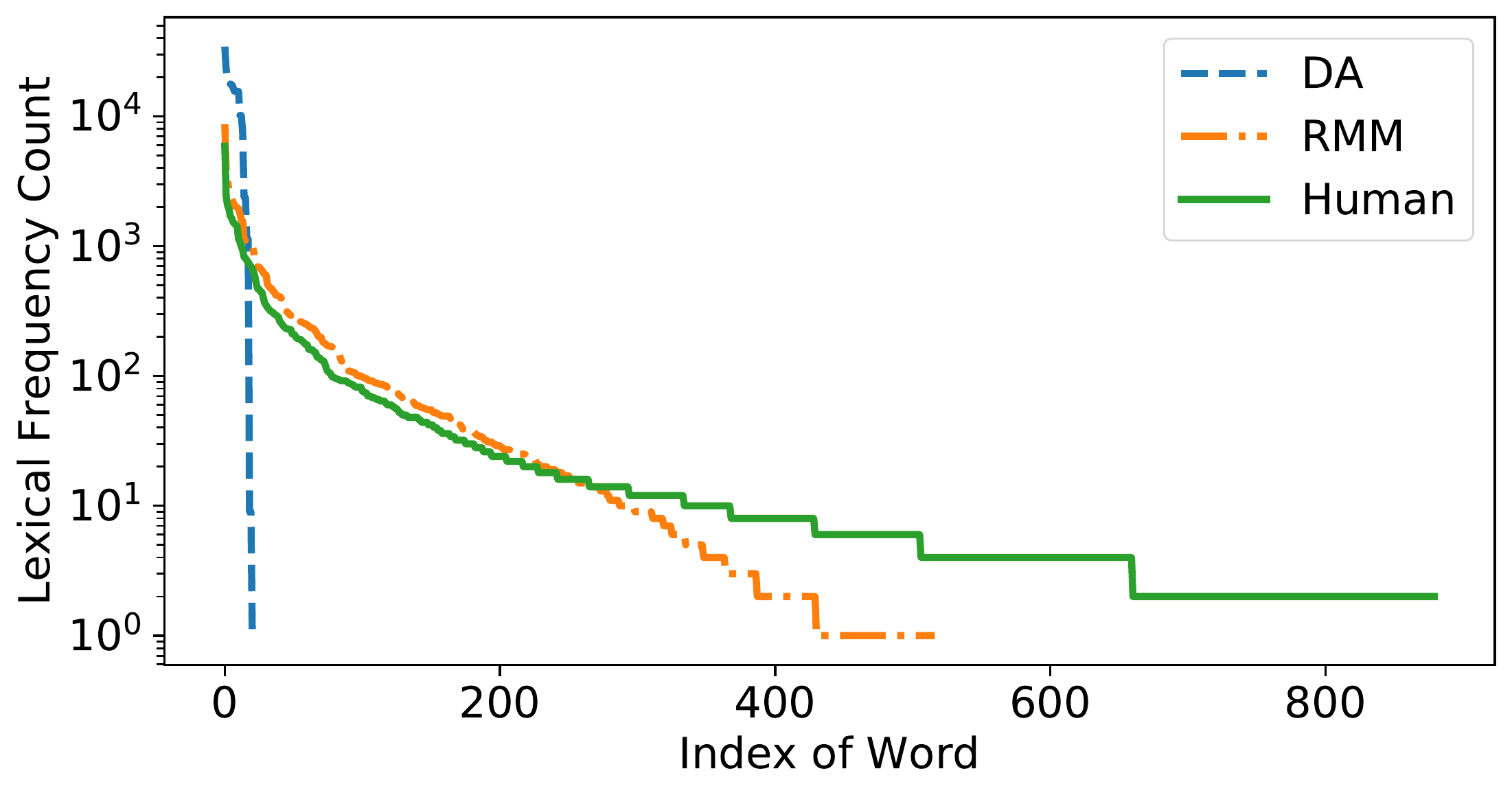}
\caption{Log-frequency of words generated by human speakers as compared to the Data Augmentation (DA) 
and our \approach{} (\acronymnormal{}) models.}
\label{fig:lexdist}
\end{figure}

%% file: 06-analysis.tex
\begin{figure*}[t!]
    \centering
    \begin{subfigure}[t]{0.47\textwidth}
        \centering
        \includegraphics[width=\textwidth]{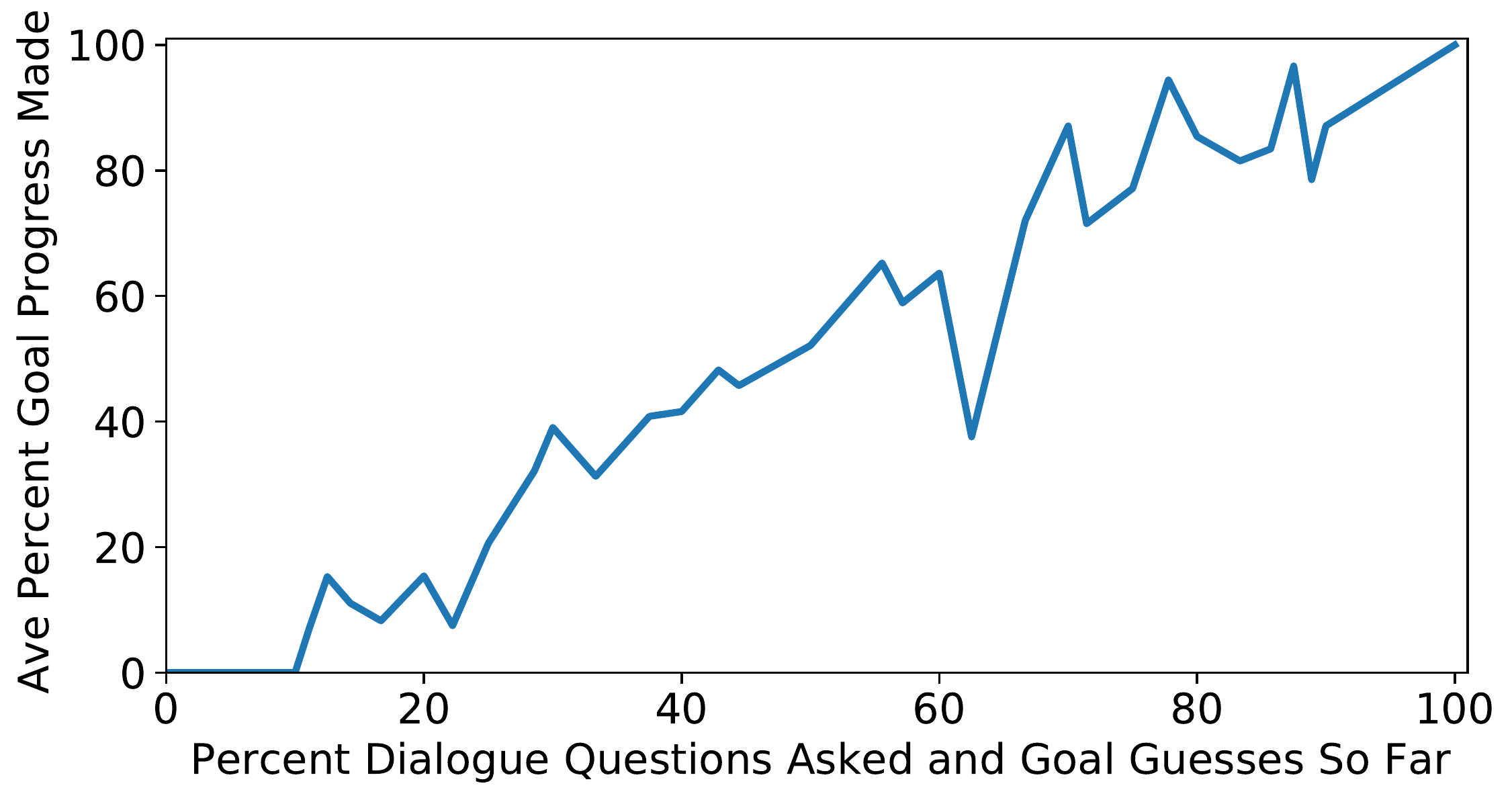}
        \caption{Human goal progress as \dia{}s unfold.
        As humans ask questions and make goal guesses, they roughly linearly make progress towards the goal location.
        }
        \label{fig:GoalQs}
    \end{subfigure}%
    \hfill 
    \begin{subfigure}[t]{0.47\textwidth}
        \centering
        \includegraphics[width=\textwidth]{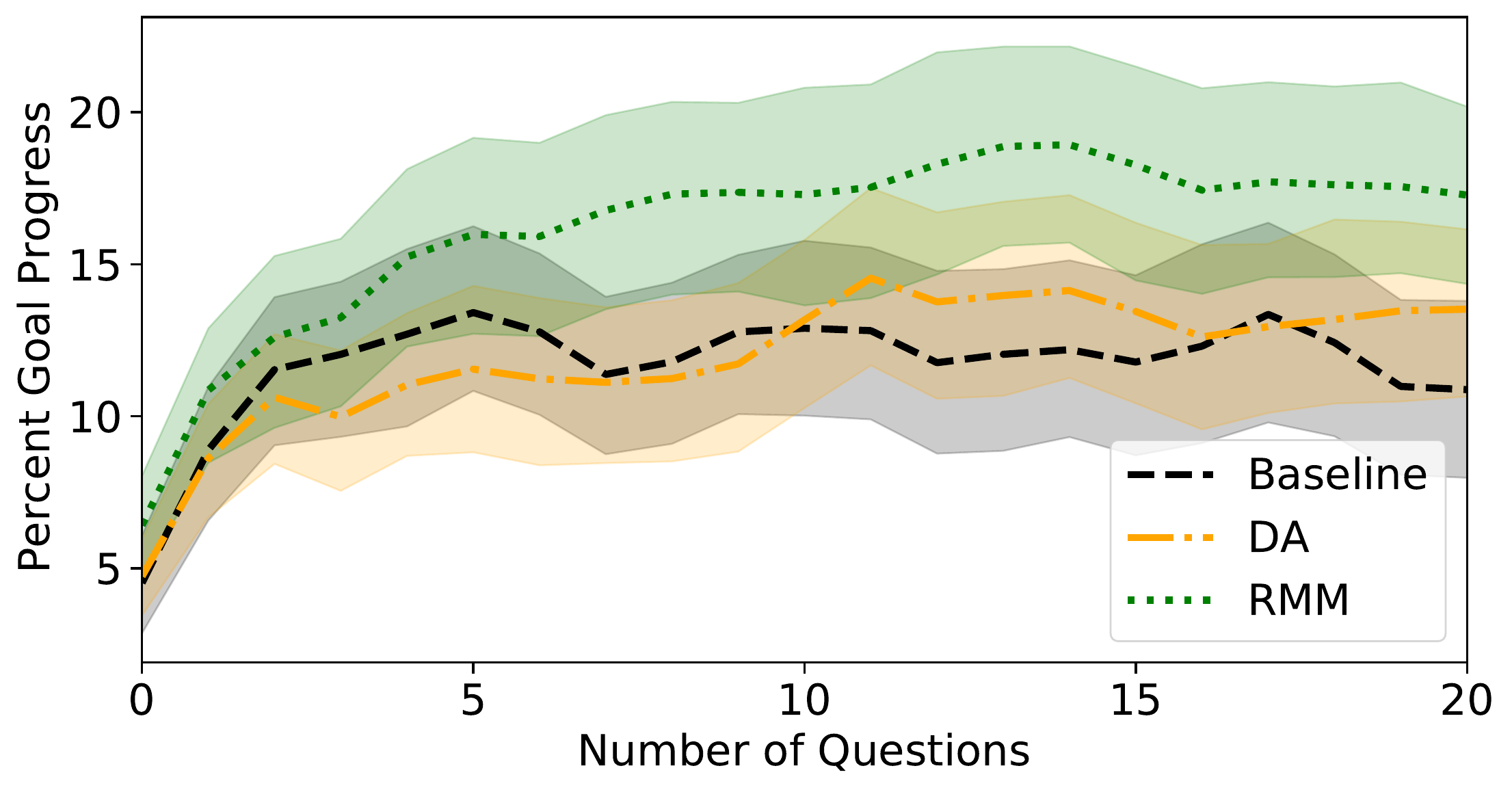} 
        \caption{Model goal progress against the number of questions.
        DA and \acronymnormal{} generated \dia{}s make slower but consistent progress (ending below 25\% of total goal progress).
        }
        \label{fig:QvGP}
    \end{subfigure}
    \caption{Effectiveness of human \dia{}s (left) versus our models (right) at reaching the goal location.
    The slopes indicate the effectiveness of each \dia{} exchange in reaching the target.}
\end{figure*}
\setlength{\tabcolsep}{1.9pt}

\section{Analysis}
\label{sec:analysis}
We analyze the lexical diversity and effectiveness of generated questions by the \acronym{}.

\subsection{Lexical Diversity}
Both \acronym{} and Data Augmentation introduce new language by exploring and the environment and generating \dia{}s.
In the case of \acronym{}, an RL loss is used to update the models based on the most successful \dia{}.
Using Data Augmentation, the best generations are simply appended to the dataset for one epoch and weighted appropriately for standard, supervised training.
The augmentation strategy leads to small boost in BLEU performance and goal progress in several settings (Table~\ref{table:cvdn_results}), but the language appears to collapse to repetitive and generic interactions.
We see this manifest rather dramatically in Figure \ref{fig:lexdist}, where the DA is limited to only 22 lexical types.
In contrast, \acronym{} continues to produce over 500 unique lexical types, much closer to the nearly 900 used by humans.

\paragraph{Human Evaluation}
We collected human judgements comparing human dialogs with generated dialogs from the baseline and RMM agents on 254 randomly selected episodes from the \textit{unseen} validation set.
While RMM uses an RL objective to inform its language generation and achieves higher progress towards the goal in this setting (Table~\ref{table:cvdn_results}), it is rated as equally or more grammatical (57\%) and as equally or more fluent (60\%) than the baseline agent, suggesting that RMM’s generated language has not devolved into a “neuralese” to achieve better task performance.
Human dialogs were rated as equally or more grammatical and fluent than RMM (89\%/83\%) and the baseline (88\%/80\%).

\subsection{Effective Questions}
The \dia{} paradigm allows us to assess the efficacy of speech acts in accomplishing goals.
In a sense, the best question elicits the answer that maximizes the progress towards the goal room.
If agents are truly effective at modeling one other, we expect the number of \dia{} acts to be minimal.

Human conversations in CVDN always reach the goal location, and usually with only 3-4 questions, as shown in Figure~\ref{fig:GoalQs}.
We see that the relationship between questions and progress is roughly linear, excusing the occasional lost and confused human teams.
The final human-human question is often simply confirmation that navigation has arrived successfully to the goal room. 

In Figure \ref{fig:QvGP}, we plot \dia{}s for the Baseline, Data Augmentation, and \acronym{} agents against percent goal progress.
The \acronym{} consistently outperforms the other two agents in terms of goal progress for each \dia{} act. We see an  increase in progress for the first 10 to 15 questions before \acronym{} levels off.
By contrast, the Baseline and Data Augmentation agents exhibit shallower curves and fail to reach the same level of performance.

\subsection{Example \Dia{}}
\begin{figure}
\centering
\includegraphics[width=0.86\columnwidth]{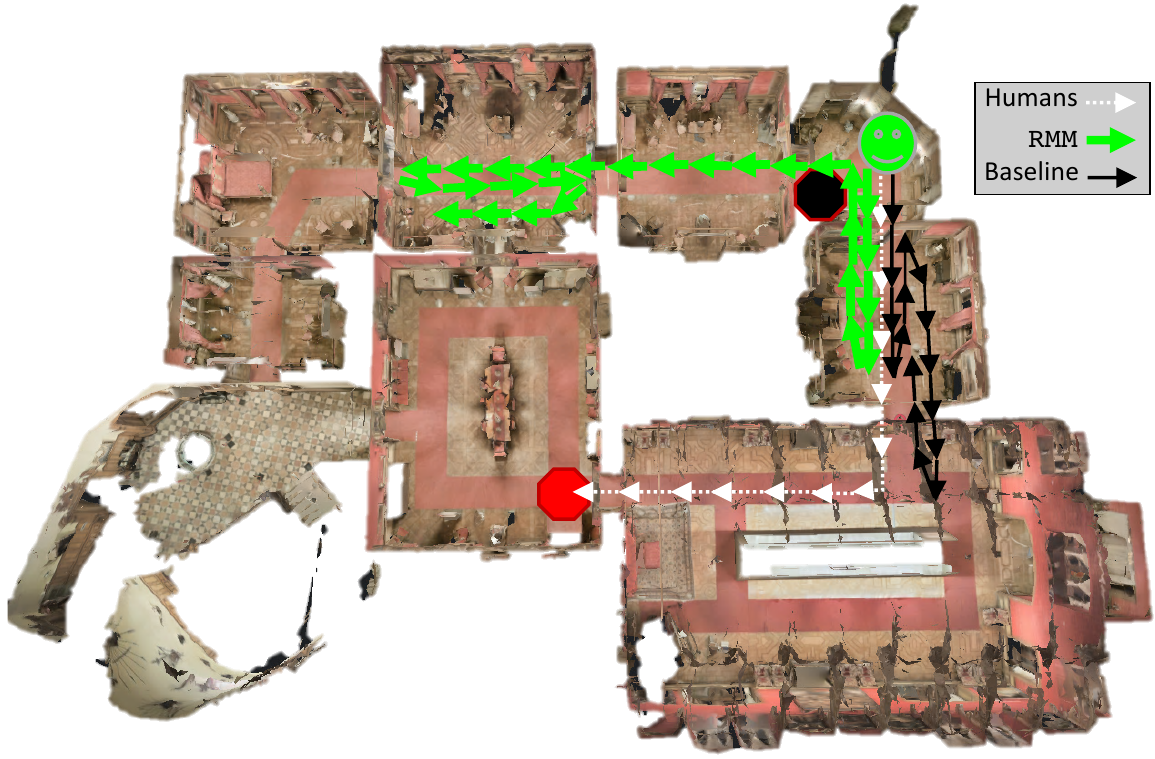}
\caption{Trajectories in an unseen environment attempting to find a target ``fire extinguisher.''
The red stop-sign is the goal room, while the black stop-sign is a non-goal room containing fire extinguishers.
The white trajectory is the human path from CVDN, black is the Baseline model, and green is our \acronym{}$_{N=3}$.}
\label{fig:val_unseen_trajectory}
\end{figure}

While Figure \ref{fig:rmm} shows a cherry-picked \acronym{} trajectory from an \textit{unseen} validation house, Figure \ref{fig:val_unseen_trajectory} gives a lemon-picked \acronym{} trajectory.
We discuss the successes and failures of a lemon-picked---showcasing model failure---trajectory in Figure~\ref{fig:val_unseen_trajectory}.
As with all CVDN instances, there are multiple target object candidates (here, ``fire extinguisher'') but only one valid goal room.
Agents can become distracted by objects of the target instance in non-goal rooms.
When the \Or{} is shown the next few shortest path steps to communicate, those steps are towards the goal room.
As can be seen in Figure \ref{fig:val_unseen_trajectory}, the learned agents have difficulty in deciding when to stop and begin retracing their steps, and in this case never arrived to the correct goal room.

The learned models' generated language is of different levels of quality, with \acronym{} language much more coherent and verbose than Data Augmentation language.
Figure~\ref{fig:conversations} shows generated conversations along with the Goal Progress (GP) at each point when a question was asked.
Note that the generation procedure for all models use the same sampler, and they start training from the same checkpoint, so the relatively coherent nature of the \acronym{} as compared to the simple repetitiveness of the Data Augmentation is entirely due to the recursive calls and RL loss.
No model uses length penalties or other generation tricks to avoid degeneration.

\begin{figure*}[ht!]
\centering
\begin{small}
\begin{tabular}{@{}l@{\hspace{5pt}}l@{\hspace{3pt}}p{7.6cm}@{\hspace{5pt}}l@{\hspace{3pt}}p{6.35cm}@{\hspace{5pt}}r@{}}
\toprule
 &  & Conversation & &&  GP\\
\midrule
\multirow{3}{*}{\rotatebox[origin=c]{90}{Human}} &\QI& Do I go in between the ropes to my right or straight forward? & \AI & straight forward through the next room  &0\\
& \QI& Should I proceed down the hall to the left of turn right? & \AI & head   down  the  hall to your  right into the  next room & 13.3 \\ 
& \QI& Should I go through the open doors that are the closest to me? & \AI & You are in  the  goal room & 29.1\\ 
\midrule
\multirow{2}{*}{\rotatebox[origin=c]{90}{DA}} 
&\QI& should i go into the room? &\AI& you are in the goal room.& 5.7\\
&\QI& should i go into the room?&\AI& you are in the goal room.& 0.0\\
\midrule
\multirow{7}{*}{\rotatebox[origin=c]{90}{\acronym{}$_{N=3}$}} 
&\QI & should i head forward or bedroom the next hallway in front of me? &\AI & yes, all the way down the small hall.	&	4.0\\
&\QI & should i turn left here?  &\AI & head into the house, then you will find a doorway at the goal staircase. go through the doors before those two small exit chairs, about half way down the hall.	&	5.7\\
&\QI & lots of sink in this house, or wrong did. ok which way do i go &\AI & go down the hallway, take a left and go down the next hallway and up the stairs on the right.	&	8.8\\
\bottomrule
\end{tabular}
\end{small}
\caption{\Dia{} samples for Figure \ref{fig:val_unseen_trajectory} with corresponding Goal Progress -- see appendix for complete outputs.
}
\label{fig:conversations}
\end{figure*}

%% file: 07-conclusions.tex
\section{Conclusions and Future Work}
We present a two-agent task paradigm for cooperative vision-and-\dia{} navigation (CVDN).
Existing work in vision-and-language navigation is largely limited to navigation only~(\ref{ch:act}), sometimes with limited additional instructions~(\ref{ch:act},\ref{ch:ans}).
By contrast, this work requires navigation~(\ref{ch:act}), question asking~(\ref{ch:que}), and question answering~(\ref{ch:ans}) components for learned, end-to-end \dia.
We find that simple speaker models 
are insufficient for the \dia{} setting, and demonstrate promising results from a recursive RL formulation with turn taking informed by theory of mind.

There are several limitations to the models presented in this paper.
We consider only agent-agent models, while the long-term goal of human-agent communication will require both human-in-the-loop training and evaluation.
Future work using \acronym{}-style modelling inspired by theory of mind will likely need to explicitly model the human interlocutor due to perceptual and communication differences~\cite{liu:aaai15}, rather than assuming the interlocutor can be modeled as a copy of oneself as in this paper.
Such modeling may incorporate world knowledge for richer notions of common ground, for example by explicitly detecting scene objects rather than using a fixed visual embedding~\cite{zhang:ijcai20}.
Additionally, we currently require the \Nr{} agent to ask questions after a fixed number of steps, while determining when to ask questions is a complex problem in itself~(\ref{ch:ask})~\cite{chi:aaai20}.
Furthermore, we use a fixed branching factor, while a dynamic branching factor in non-parametric learning setting can incorporate the uncertainty of the policy model. 

We hope this task paradigm will inspire future research on learning agent-agent, task-oriented communication with an eye towards human-agent cooperation and language-guided robots.

%% file: Appendix.tex
\clearpage
\onecolumn 
\section{Appendix}

\subsection{Additional Reproducibility details}
\begin{enumerate}
    \item Hardware:  Single NVIDIA P100 GPU
    \item Training times:\\
    \begin{tabular}{lcr}
        \toprule
        Setting     & Iterations & Average time\\
        \midrule
        Pretraining & 20K & 273m\\
        Baseline    & 5k & 1,622m\\
        Data Aug    & 5k & 1,161m\\
        RMM1        & 5k & 4,205m\\
        RMM3        & 5k & 6,590m\\
        \bottomrule
    \end{tabular}
    \item Model parameters:\\ 
        \begin{tabular}{ll}
        Speaker & 3.5M\\
        Action Decoding & 4.7M\\
        \end{tabular}
    \item Hyperparameters:\\
    \begin{tabular}{ll}
        Bounds for temperature sampling& [0.1-2.0]\\
        Bounds for lambda DA contribution& [0.1-1.0]\\
        Trials for temperature sampling& [0.1, 1.0, 2.0]\\
        Trials for lambda DA contribution& [0.1, 0.25, 0.5, 0.75, 1.0]\\
        Method for choice& Grid search\\
    \end{tabular}
\end{enumerate}

\subsection{Human Evaluation details}
The table below shows the full results of the human evaluation on a randomly selected subset of 245 \textit{unseen} environment dialogues.
Questions asked:
\vspace{10pt}
\begin{enumerate}[noitemsep,nolistsep,label={\bfseries Q\arabic*}]
\item \label{q:gram} Which dialog is more grammatically correct?
\item \label{q:flue} Which dialog is more fluent?
\item \label{q:goal} Do the two dialogues seem to have the same goal?
\end{enumerate}
\vspace{15pt}
\begin{tabular}{lcclc}
    & \multicolumn{2}{c}{\bf \% Favored} & \multicolumn{2}{c}{\bf \% Same Goal} \\
    \bf Model & \ref{q:gram} & \ref{q:flue} & & \ref{q:goal} \\
    \toprule
    RMM & 35.0 & 37.0 & Same & 49.6 \\
    Baseline & 42.5 & 39.8 & Diff & 50.4  \\
    No Difference & 22.4 & 23.2 & & \\
    \midrule
    RMM & 11.0 & 17.3 & Same & 44.9 \\
    Human & 76.8 & 67.7 & Diff & 55.1 \\
    No Difference & 12.2 & 15.0 & & \\
    \midrule
    Baseline & 12.2 & 20.1 & Same & 45.7 \\
    Human & 78.0 & 63.8 & Diff & 54.3 \\
    No Difference & \phantom{0}9.8 & 16.1 & & \\
    \bottomrule
\end{tabular}

\clearpage

\subsection{Conversation Examples}

Full human and baseline conversations for Figure \ref{fig:val_unseen_trajectory}.

\noindent
\begin{small}
\begin{tabular}{@{}l@{\hspace{5pt}}l@{\hspace{3pt}}p{14cm}@{\hspace{5pt}}r@{}}
\toprule
 &  & Conversation & GP\\
\midrule
\multirow{6}{*}{\rotatebox[origin=c]{90}{Human}} &\QI& Do I go in between the ropes to my right or straight forward? &0\\
& \AI& straight forward through the next room & \\
& \QI& Should I proceed down the hall to the left of turn right? & 13.31 \\ 
& \AI& head   down  the  hall to your  right into the  next room & \\
& \QI& Should I go through the open doors that are the closest to me? & 29.05\\ 
& \AI& You are in  the  goal room & \\
\midrule
\multirow{12}{*}{\rotatebox[origin=c]{90}{Baseline}} 
& \QI & ok now which way ?	& 4.0\\
& \AI & keep going forward and then into a bedroom on your left . it ' s to the left of the bed .	 & \\
& \QI & do i continue going to the right into the living room ?	& 5.7 \\
& \AI & go into the door in front of you , then an right , the bathroom ' s there .	 & \\
& \QI & do i keep going to the right into the living room ?	& 8.8\\
& \AI & it looks like that is the goal room .	 & \\
& \QI & should i go into the laundry room to my right , or do i need to go in ?	& 7.5 \\
& \AI & i see many ways to tell . you just go around . you ' re in the right room .	 & \\
& \QI & ok now which way ?	& 4.0 \\
& \AI & go into the room near you . that should be the goal	 & \\
& \QI & okay i see any set of stairs . should i go up them ?	& 0.4 \\
& \AI & turn right , then another right , the bathroom ' s there .	 & \\
& \QI & ok . can you go more directions . i see many rooms on this floor .	& 0 \\
& \AI & turn right , then another right , the bathroom ' s there .	 & \\
& \QI & should i go the room ? or into the door to my left ?	& 4.0 \\
& \AI & turn right , then another right , the bathroom ' s there .	 & \\
& \QI & should i stay in this room or go out ?	& 8.8 \\
& \AI & turn right , then another right , the bathroom ' s there .	 & \\
& \QI & ok , should i go in this direction ?	& 9.9 \\
& \AI & turn right , go past the pool table , into the room just towards the stairs . that should be the goal .	 & \\
& \QI & should i go left toward the room , or past the painting and straight ? or to my right ?	& 7.5 \\
& \AI & turn right , then another right , the bathroom ' s there .	 & \\
& \QI & should i take a left here in the room ?	& 5.7 \\
& \AI & that is the goal room	 & \\
& \QI & should i stay in this room or go out ?	& 8.8 \\
& \AI & you see the front , that is the room . its goal room	 & \\
& \QI & the room ? or into the door to my left ?	& 13.3 \\
& \AI & go up the stairs in front of you , then to your left , you ' ll see a bathroom . i think that bathroom is the goal	 & \\
& \QI & should i go left toward the room , or past the painting and straight ? or to my right ?	& 9.3 \\
& \AI & go towards the table in front of me of the hallway after take a right , then make a left into the sitting room .	 & \\
& \QI & should i go to the right towards the blue painting ?	& 9.3 \\
& \AI & continue going up , go to the bedroom	 & \\
& \QI & should i go into the door on the left , or keep going forward ?	& 11.1 \\
& \AI & go into the bedroom near you . that should be the goal	 & \\
& \QI & do i need to go up ?	& 8.8 \\
& \AI & go out in the room , make a right past the blue painting and go up the stairs	 & \\
& \QI & ok now which way ?	& 7.5 \\
& \AI & go up the spiral stairs again when you ' go into that bedroom straight ahead .	 & \\
& \QI & should i go outside ?	& 8.8 \\
& \AI & a bathroom on pretty left , go in it and then into the bathroom in that bedroom	 & \\
\midrule
\setlength{\tabcolsep}{1.4pt}
\end{tabular}
\end{small}

\clearpage

Data Augmentation and \acronym{} conversations for Figure \ref{fig:val_unseen_trajectory}.

\noindent
\begin{small}
\begin{tabular}{@{}l@{\hspace{5pt}}l@{\hspace{3pt}}p{14cm}@{\hspace{5pt}}r@{}}
\toprule
 &  & Conversation & GP\\
 \toprule
 \multirow{4}{*}{\rotatebox[origin=c]{90}{DA}} 
&\QI& should i go into the room?& 5.66\\ 
&\AI& you are in the goal room.&\\
&\QI& should i go into the room?& 0.03\\
&\AI& you are in the goal room.&\\
& & \dots{} \textit{(repeats until maximum trajectory length is reached)} & \\
\midrule
\multirow{6}{*}{\rotatebox[origin=c]{90}{\acronym}} 
&\QI & should i head forward or bedroom the next hallway in front of me ?	&	4.0\\
&\AI & yes , all the way down the small hall .	&	\\
&\QI & should i turn left here ?	&	5.7\\
&\AI & head into the house , then you will find a doorway at the goal staircase . go through the doors before those two small exit chairs , about half way down the hall .	&	\\
&\QI & lots of sink in this house , or wrong did . ok which way do i go	&	8.8\\
&\AI & go down the hallway , take a left and go down the next hallway and up the stairs on the right .	&	\\
&\QI & right place now ?	&	7.5\\
&\AI & turn left and when you reach to the bathroom . before these go up the stairs on your left , and walk passed the mirror and chairs on the left into hallway across from the stairs	&	\\
&\QI & do i go left , right , or up the stairs ?	&	4.0 \\
&\AI & turn to your left and keep going straight	&	\\
&\QI & do i make a right turn or shall i enter the room to my left ?	&	0.4\\
&\AI & turn left to go down the hall , then that fire bed has that	&	\\
&\QI & left or continue ?	&	0\\
&\AI & yes , go past the dining table and take an immediate right . head through the small door to the left of the window before those two way doors behind you , go up the second small set of stairs .	&	\\
&\QI & should i go downstairs or turn left ?	&	4.0\\
&\AI & go to the right and go to the front doorway .	&	\\
&\QI & should i go down the stairs or stay on this floor ?	&	8.8\\
&\AI & take a left towards the bathroom , then take a left before it and go all the way down the hall	&	\\
&\QI & do i go up these is to the right or right of the steps ?	&	9.9\\
&\AI & go to the left side of the staircase and turn left in the doorway before the two small office chairs , about half way down the hall .	&	\\
&\QI & should i turn left , go straight into the living room , or up the stairs ?	&	7.5\\
&\AI & turn to your right and go straight down the hall	&	\\
&\QI & do i go out into the hallway ?	&	5.7\\
&\AI & go left down the hall where the office floor . and pass the second door after a right and table .	&	\\
&\QI & ok , should i go right or left next ?	&	8.8\\
&\AI & go back to the staircase . go through the doorway you and before the hallway on the right .	&	\\
&\QI & do i make a left turn or shall i enter the room to my left ?	&	13.3\\
&\AI & go down the hall and turn right into the bedroom	&	\\
 &\QI & should i go to the left or the right ?	&	9.3\\
 &\AI & yes , go out of this room , turn right and go down the white hall before the staircase stairs , then go down the way down that way you get .	&	\\
 &\QI & ok i was a in by this office painting , or i just in the second hallway in front of me ?	&	9.3\\
 &\AI & okay .	&	\\
 &\QI & which way do i go in , or do i head up the stairs ?	&	11.1\\
 &\AI & go all the way to the one of the staircase . turn left in the doorway before the two two office chairs , about half way down the hall .	&	\\
 &\QI & ok wrong far which way do i go	&	8.8\\
 &\AI & right then at the top of the stairs .	&	\\
 &\QI & left or continue ?	&	7.5\\
 &\AI & yes . go down the hall and stop at the landing of the stairs .	&	\\
\bottomrule
\end{tabular}
\end{small}
%